\RequirePackage{fix-cm}
\documentclass[twocolumn]{svjour3}          
\smartqed  
\usepackage{times}
\usepackage{epsfig}
\usepackage{amssymb}
\usepackage{siunitx}
\usepackage{diagbox}

\usepackage{graphicx}

\usepackage{natbib}
\usepackage{amsmath}
\usepackage{multirow}
\usepackage{color}
\usepackage{amsfonts}

\usepackage[colorlinks,citecolor=blue]{hyperref}

\usepackage{subcaption}

\usepackage{float}

\begin{document}

\title{LAMP-HQ: A Large-Scale Multi-Pose High-Quality Database and Benchmark for NIR-VIS Face Recognition
}


\author{Aijing Yu\and Haoxue Wu\and Huaibo Huang\and Zhen Lei\and Ran He}
 \institute{Aijing Yu, Huaibo Huang, Ran He   \\
           {$^1$}Center for Research on Intelligent Perception and Computing, CASIA, Beijing, China\\
           {$^2$}National Laboratory of Pattern Recognition, CASIA, Beijing, China\\
           {$^3$}School of Artificial Intelligence, University of Chinese Academy of Sciences, Beijing, China \\
           \email{\{huaibo.huang, aijing.yu\}@cripac.ia.ac.cn, \\
            rhe@nlpr.ia.ac.cn} \\
           \and
           Haoxue Wu, Zhen Lei\\
           {$^1$}Center for Biometrics and Security Research, CASIA, Beijing, China\\
           {$^2$}National Laboratory of Pattern Recognition, CASIA, Beijing, China\\
           {$^3$}School of Artificial Intelligence, University of Chinese Academy of Sciences, Beijing, China \\
           \email{wuhaoxue2019@ia.ac.cn, \\zlei@nlpr.ia.ac.cn}
}

\date{Received: date / Accepted: date}
\maketitle

\begin{abstract}
Near-infrared-visible (NIR-VIS) heterogeneous face recognition  matches NIR to corresponding VIS face images. However, due to the sensing gap, NIR images often lose some identity information so that the recognition issue is more difficult than conventional VIS face recognition. Recently, NIR-VIS heterogeneous face recognition has attracted considerable attention in the computer vision community because of its convenience and adaptability in practical applications.  Various deep learning-based methods have been proposed and substantially increased the recognition performance, but the lack of NIR-VIS training samples leads to the difficulty of the model training process. In this paper, we propose a new $\textbf{L}\textbf{a}$rge-Scale  $\textbf{M}$ulti-$\textbf{P}$ose $\textbf{H}$igh-$\textbf{Q}$uality NIR-VIS database ‘$\textbf{LAMP-HQ}$’ containing 56,788 NIR and 16,828 VIS images of 573 subjects with large diversities in pose, illumination, attribute, scene and accessory. We furnish a benchmark along with the protocol for NIR-VIS face recognition via generation on LAMP-HQ, including Pixel2-\\Pixel, CycleGAN, and ADFL. Furthermore, we propose a novel exemplar-based variational spectral attention network to produce high-fidelity VIS images from NIR data. A spectral conditional attention module is introduced to reduce the domain gap between NIR and VIS data and then improve the performance of NIR-VIS heterogeneous face recognition on various databases including the LAMP-HQ.
\end{abstract}

\section{Introduction}
\label{intro}

\begin{figure}[htb]
\begin{center}
\includegraphics[width=0.95\linewidth]{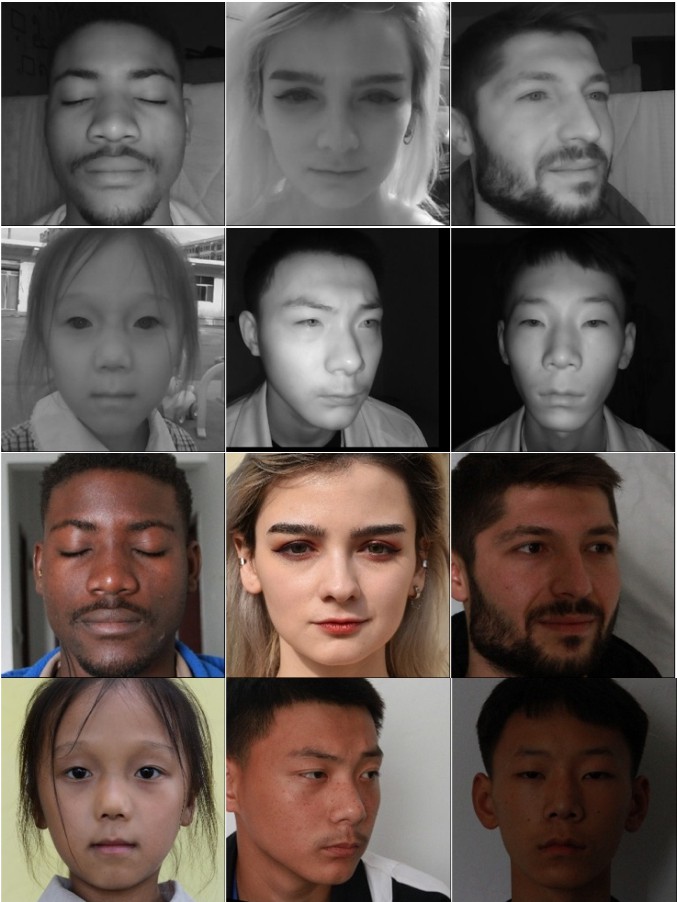}
\end{center}
\vspace{-0.1cm}
   \caption{Examples in the LAMP-HQ database. The top two rows are NIR images, and the bottom two are VIS images of the same identities. }
   \label{figure:database}
   \vspace{-0.3cm}
\end{figure}


Conventional face recognition under controlled visual light has been one of the most studied directions in the community of computer vision in
recent years~\citep{ouyang2016survey}. It has been applied in various fields, even achieving better performance than humans in most cases. Recently, more attention has been focus on heterogeneous recognition issue, such as sketch to photo~\citep{tang2002face,wang2014comprehensive}, near-infrared to visible~\citep{li2013casia,xiao2013coupled}, polarimetric thermal to visible~\citep{di2018polarimetric}, and cross resolutions~\citep{biswas2011multidimensional}. Due to the insensitivity to illumination~\citep{zhu2014matching}, near-infrared (NIR) devices are widely used in monitoring and security systems. This leads to the issue of NIR-VIS heterogeneous face recognition (HFR), where the NIR images captured under near-infrared lighting are always matched to the registered VIS images. Because it is harder to match face images across different spectral, massive efforts have been invested in the community to address this issue~\citep{reale2016seeing,he2018wasserstein}. With the development of deep neural network, several deep learning-based methods~\citep{reale2016seeing,saxena2016heterogeneous,wu2019disentangled} have been suggested to improve the performance of HFR.

However, challenges still exist in the following three aspects: {\bf(1) Sensing gap.} NIR and VIS face images are captured under various  illuminations by different devices, which leads to different textural and geometric appearances between images in probe and gallery. Therefore, it is ineffective to some degree to directly employ traditional face recognition methods in NIR-VIS face recognition ~\citep{reale2016seeing,saxena2016heterogeneous}. {\bf(2) Pose variations.} Most NIR face images captured by NIR sensors contain various poses with diverse angles, expressions and accessories, leading to incomplete face information. However, VIS faces are often frontal in the corresponding recognition database. Large discrepancies result in obstacles to the process of recognition\citep{li2006highly,li2006near}. {\bf(3) Small-scale dataset.} Due to the widespread use of the Internet, it is relatively easy to obtain a great collection of VIS face images. However, NIR face images are often captured by special NIR sensors so that it is still expensive and time consuming to collect a large-scale NIR-VIS face image database.

\begin{table*}[htb]
\begin{center}
\label{tab:comparison}
\caption{Comparisons with existing NIR-VIS facial databases.}
\vspace{0.2cm}
\begin{tabular}{l|ccccc}
\hline
Database    & No. of Image & NIR Image size & VIS Image size & Subjects & Year \\
\hline
NIR-VIS 2.0 & 17,580        & $640\times480$       & $640\times480$        & 725     & 2013 \\
BUAA        & 2700         & /              & /              & 150      & 2012 \\
Oulu        & 7,680        & /              & /              & 80       & 2009 \\
\hline
LAMP-HQ        & 73,616        & $960\times720$        & $5184\times3456$      & 573      & 2020\\
\hline
\end{tabular}
\end{center}
\end{table*}

In this paper, we propose a LAMP-HQ database to alleviate the issues mentioned above. The comparisons with existing other databases are summed up in Table 1. The main advantages of our new database lie in the following aspects: {\bf(1) Large-scale.} LAMP-HQ contains 56,788 NIR and 16,828 VIS images of 573 subjects  with 3 races (containing Asian, white and black), 3 expressions, 5 scenes, 3 angles, broad age distribution (ranging from 6 to 70) and different accessories. {\bf(2) Multi-scenes.} Different from previous databases, we collect face images under 5 illumination scenes, including indoor natural light, indoor strong light (with fluorescent), indoor dim light (drawing the curtains), outdoor natural light and outdoor backlight. {\bf(3) Multi-poses.} We capture VIS and NIR images with 3 yaw angles ($\ang{0}$,$\pm\ang{45}$); on this basis, we capture more images from the side and bottom view, especially to NIR. We also acquire the solution of closing eyes and smiling (including grin and smile). {\bf(4) High-resolutions.} VIS images are captured by Canon-7D, and NIR images are captured by AuthenMetric-CE31SE, which leads to high-resolution images ($5184\times3456$ and $960\times720$ , respectively). {\bf(5) Accessory.} We employ 15 types of glasses as accessories to rich the diversity of our database. Headdresses and earrings are also preserved to further increase the complexity of the database. To the best of our knowledge, LAMP-HQ is the largest-scale NIR-VIS database containing various races, ages, angles, expressions, scenes, illuminations and accessories. We will release the new database in the near future. In addition, we provide an effective benchmark on a few state-of-the-art methods, including Pixel2Pixel~\citep{isola2017image}, CycleGAN~\citep{zhu2017unpaired}, and ADFL~\citep{song2018adversarial}.

To further address the challenges in HFR, we propose a novel exemplar-based variational spectral attention network (VSANet) to transfer NIR images to VIS images that are more efficient for recognition.
As illustrated in Fig.~\ref{fig:framework}, VSANet contains three sub-modules, i.e., a spectral variational autoencoder (SVAE), a cross-spectral UNet (CSU), and a spectral conditional attention (SCA) module. The spectral variational autoencoder models the spectral style of VIS data using a variational representation that approximately matches a prior distribution. The cross-spectral UNet is utilized to generate high-fidelity VIS images from the input NIR data along with the referenced spectral style.  In addition, a spectral conditional attention module is presented to accurately transfer both the global and local spectral information accurately, where the global and local spectral styles are represented by the extracted latent and feature maps of the middle layers in SVAE.
For spectral styles that can be sampled from the posterior of VIS data or the prior, the proposed method is capable of producing a VIS image from an NIR input with or without a reference VIS image.

The contributions of our paper lie in 3-fold:

1) We release a new LAMP-HQ database to add to the research progress of HFR. The database contains 56,788 NIR and 16,828 VIS images of 573 subjects with a wide variety of poses, illumination, attributes, environments and accessories. We believe that the new LAMP-HQ database can significantly  advance NIR-VIS face analysis, similar to a lamp lighting up the dark.

2) We provide a comprehensive qualitative and quantitative efficient benchmark of several state-of-the-art methods for NIR-VIS heterogeneous face recognition (HFR), including Pixel2Pixel~\citep{isola2017image}, CycleGAN~\citep{zhu2017unpaired}, and ADFL~\citep{song2018adversarial}. The performance on the LAMP-HQ reflects its challenges and difficulties.

3) We propose a novel exemplar-based variational spectral attention network (VSANet), including three modules to learn and transfer the spectral style to the generated VIS data from the NIR input. The spectral conditional attention mechanism makes it able to guide the generation of VIS data using both global and local spectral information.

\section{Related Work}
\label{Rel}
\subsection{NIR-VIS Databases}

\label{nir:2}
There are three most commonly used NIR-VIS databases to evaluate the recognition performance in the community. {\bf(1) The CASIA NIR-VIS 2.0 Face Database}~\citep{li2013casia} is the largest and most challenging public NIR-VIS database in existence. It consists of 725 subjects, each of which has 1-22 VIS and 5-50 NIR images with large diversities in illumination, expression, distance, and pose. Each image is randomly captured, so the NIR and VIS images of one person are unpaired. There are two views of protocols designed in the database: one is adopted to fine-tune the super-parameter and the other is utilized in the process of training and testing. The protocol in the View 2 includes 10-fold experiments. There are approximately 2,500 VIS and 6,100 NIR images of approximately 360 subjects in the training fold and the other images from the remaining 358 subjects are employed for the testing. {\bf(2) The BUAA-VisNir face database}~\citep{huang2012BUAA} is often used to test the performance of domain adaptation~\citep{shao2014generalized}. It contains 150 subjects with 9 paired VIS and NIS images synchronously captured by a single multi-spectral camera. All images of every subject are collected under 9 poses and expressions (including neutral-frontal, left-rotation, right-rotation, tilt-up, tilt-down, happiness, anger, sorrow and surprise). The training set includes 900 images of 50 subjects, and the testing set contains the remaining 1,800 images of 100 subjects. {\bf(3) The Oulu-CASIA NIR-VIS database}~\citep{chen2009learning} includes 80 identities, 30 of them are from the CASIA, and the rest are from the Oulu University. All the NIS and VIS images are collected under three types of illumination environments (normal indoor, weak and dark) and all the subjects are shot with six various expressions (anger, disgust, fear, happiness, sadness and surprise). Based on the protocol designed in~\citep{shao2016cross}, 10 subjects of the Oulu University and 30 subjects of the CASIA are singled out to build a database. Eight images from each expression are selected at random so that there are 48 NIR and 48 vis images.

\subsection{Heterogeneous Face Recognition}
Heterogeneous face recognition (HFR) problem has attracted increasing attention in the recent years~\citep{sarfraz2015deep,riggan2016estimation,liu2016transferring,zhang2018dual,di2019polarimetric}. NIR-VIS face recognition has been one of the most representative and studied issues in the research field. 
In this section, we review various recent advances of HFR from three aspects~\citep{ouyang2016survey,zhu2017unpaired}: image synthesis, latent subspace, and domain-invariant features.

{\bf Image synthesis} methods synthesize images from one domain to another and then match the heterogeneous images in the same spectral domain. Method in~\citep{tang2003face} firstly attempts to synthesize a sketch photo from a visual face image. Markov random fields are adopted in~\citep{wang2008face}, which could transfer the sketch to face photo in a multi-scale way. Lei~et al.~\citeyearpar{lei2008face} utilize canonical correlation analysis (CCA) to synthesize 3D face from the NIR one. Work~\citep{wang2009analysis} designs an analysis-by-synthesis framework for heterogeneous face mapping. The dictionary learning is employed in~\citep{wang2012semi,huang2013coupled,juefei2015nir} during the process of reconstruction. Zhang~et al.~\citeyearpar{zhang2018dual} propose a dual-transfer synthesis framework which divide the transfer process into the intra- and inter- domain.

Benefiting from the essential advanced in deep learning, recent there are various impressive works based on deep networks. A VIS feature estimation and VIS image reconstruction two-step procedure is introduced in~\citep{riggan2016estimation} to transfer polarimetric thermal images to visual domain. Lezama~et al.~\citeyearpar{lezama2017not} introduce a cross-spectral hallucination and low-rank embedding to generate heterogeneous images in a patch-based way. Method in~\citep{di2018polarimetric}  employs the attributes extracted from the visible image in the synthesis progress. Work~\citep{huang2017beyond} utilize a global and local perception GAN~\citep{goodfellow2014generative} to deal with the face rotation issue. Song~et al.~\citeyearpar{song2018adversarial} utilize a Cycle-GAN\citep{zhu2017unpaired} to integrate cross-spectral face hallucination and discriminative feature learning on both raw-pixel space and compact feature space and then improve the performance of HFR. Zhang et al.~\citeyearpar{zhang2019synthesis} employ the GAN model to synthesize visual images from polarimetric thermal domain. A self-attention mechanism is used in~\citep{di2019polarimetric} to guide the generating of visual images.

{\bf Latent subspace} methods aim to map images of two different domains into a common latent, where the features of heterogeneous images can be matched. Lin et al.~\citeyearpar{lin2006inter} propose the Common Discriminant Feature Extraction (CDFE) to extract the discriminant and locality feature. The coupled discriminant analysis is proposed in~\citep{lei2012coupled} followed by~\citep{huang2012regularized}, which brings up a regularized discriminative spectral regression method. Wang et al.~\citeyearpar{wang2015joint} propose a baseline method of NIR-VIS HFR by adopting the feature selection. Yi~et al.~\citeyearpar{yi2015shared} employ Restricted Boltzmann Machines to learn a locally shared feature to reduce the heterogeneity around every facial point. Work~\citep{kan2015multi} proposes the multi-view discriminant analysis to reduce the domain discrepancy. In order to preserve the common information of images in different domains, work~\citep{li2016mutual} adds the mutual component analysis to the process of mapping.
Jin et al.~\citeyearpar{jin2017multi} use the extreme learning machine (ELM) integrated with the multi-task clustering for cross-spectral feature learning. He~et al.~\citeyearpar{he2018wasserstein} fabricate a hierarchical network to learn both modality-invariant feature subspace and modality-related spectrum subspace. An orthogonal dictionary alignment is proposed in~\citep{mudunuri2018dictionary} to deal with the pose and low-resolution issues of NIR images.

{\bf domain-invariant feature} methods extract the features related to common identity information of heterogeneous face images. In traditional methods, hand-crafted feature descriptors are utilized, i.e. Local Binary Patterns (LBP), (Scale-invariant Feature Transform) SIFT, Histograms of Oriented Gradients (HOG), Gabor filters, and Difference of Gaussian (DoG)~\citep{klare2010matching,goswami2011evaluation,klare2012heterogeneous,zhu2014matching}. Zhang et al.~\citeyearpar{zhang2011coupled} introduce a novel feature information based on mutual information of heterogeneous images. Huang et al.~\citeyearpar{huang2012learning} use a pair of heterogeneous face databases as generic training databases, and find the common local geometrical information of heterogeneous face samples and generic training samples. Gong et al.~\citeyearpar{gong2017heterogeneous} transform the facial pixels into an encoded face space, where the features could be matched directly.

In addition, deep learning based schemes have been recently developed. Liu et al.~\citeyearpar{liu2016transferring} utilize the ordinal activation function to select discriminative features and employ two types of triple loss in the process of transform. ~\cite{saxena2016heterogeneous} discusses various metric learning strategies to increase the HFR performance based on the pre-trained VIS CNN. Work~\citep{reale2016seeing} suggests using deeper networks to learn the features of cross-modal face images and proposes two novel network structures with small convolutional filters. ~\cite{he2017learning} employs a two-level CNN to learn domain-invariant identity representation and modality-related spectrum representation. Sarfraz and Stiefelhagen~\citeyearpar{sarfraz2017deep} exploit a deep neural network to learn a non-linear mapping of images in different domains and preserve the identity information meanwhile. To learn information from raw facial patches directly, Peng et al.~\citeyearpar{peng2019dlface} propose a deep local descriptor learning framework with a novel cross-modality enumeration loss.

\section{Database Description}
\label{dat}
In this section, we introduce the database at length, including the process of data collection and data cleaning. A training and evaluation protocol is defined at the end of this section.
\begin{figure}[htb]
\begin{center}
\includegraphics[width=0.95\linewidth]{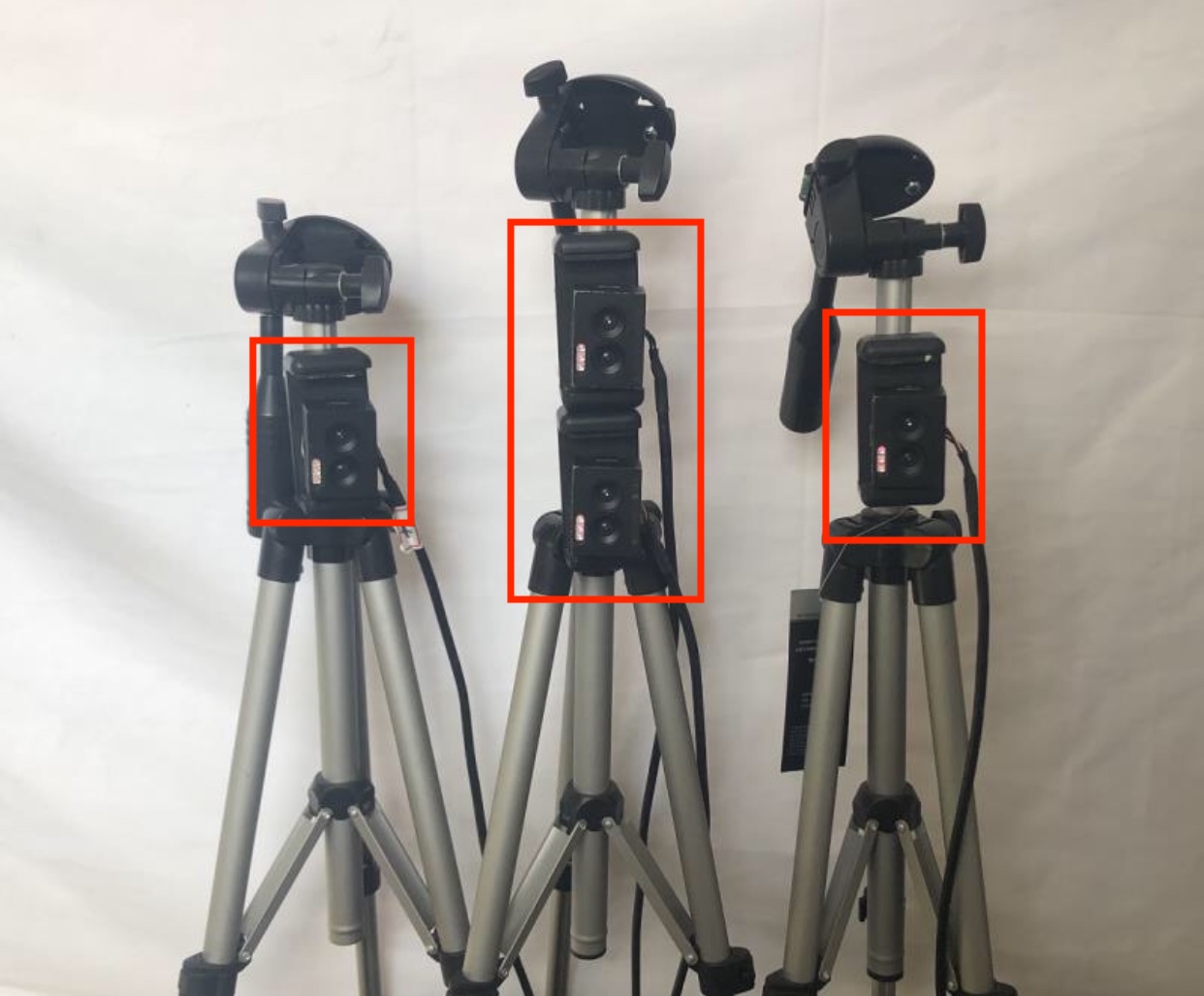}
\end{center}
\vspace{-0.1cm}
   \caption{The capture device of NIR images. The cameras are in the red boxes.}
   \label{figure:device}
   \vspace{-0.3cm}
\end{figure}
\subsection{Data Collection and Cleaning}
\begin{figure*}[htb]
\begin{center}
\includegraphics[width=0.85\linewidth]{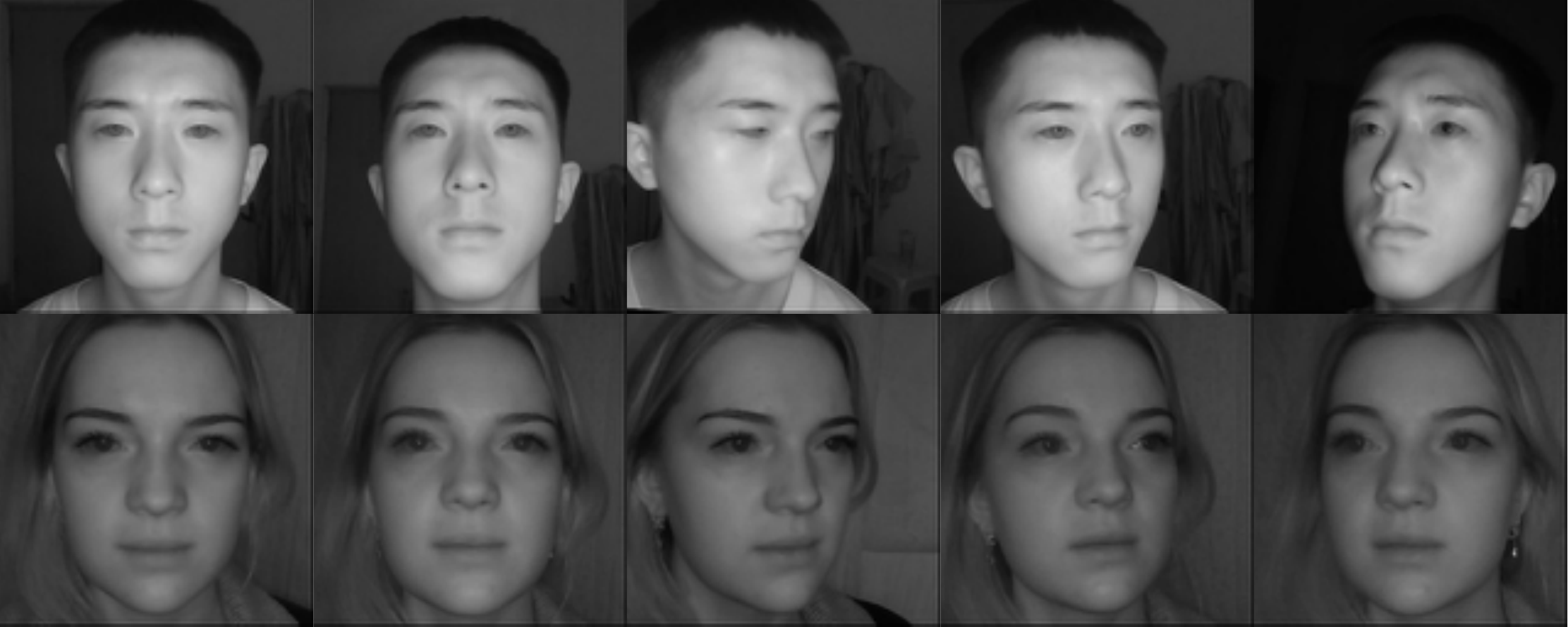}
\end{center}
\vspace{-0.1cm}
   \caption{Examples of different poses of NIR images in LAMP-HQ.}\label{figure:multipose}
   \vspace{-0.3cm}
\end{figure*}

We use Canon-7D and AuthenMetric-CE31S to acquire VIS and NIR images respectively. We prepare 5 types of illumination scenes during shooting, including indoor natural light, indoor strong light (with fluorescent), indoor dim light (drawing the curtains), outdoor natural light and outdoor backlight. The outdoor scenes will be reduced to one type when raining. Each subject has 6 different pictures, containing 3 expressions (neutral, closing eyes and smile or grin), 1 accessory (glasses) and 3 poses with 3 yaw angles ($\ang{0}$,$\pm\ang{45}$) in per illumination scene. Some sample images are showed in Fig.~\ref{figure:database}. In addition, to advance  NIR-VIS face recognition in the wild, we design a device to capture more side- and bottom-viewed NIR data, which is closer to  real-world application. The detailed collecting device is shown in the Fig.~\ref{figure:device}. Fig.~\ref{figure:multipose} shows some examples of these views. The photographic distance and height are not strictly regulated to increase the complexity of the database. Finally, we capture $573 \times 6 \times 5$ = 17,190 ($p \times a \times i$) VIS and $573 \times 6 \times 5 \times 4$ = 68,760 ($p \times a\times i \times l$) NIR face images, where $p$, $a$, $i$ and $l$ denote the number of participants, attributes, illuminations and  lens. Then, we manually check all the facial images and remove those blurred or incomplete images. Finally, we preserve 16,828 VIS and 56,788 NIR images of 573 subjects, each subject contains 29 VIS and 99 NIR images on average.

\begin{table}[htb]
\centering
\label{tab:attributes}
\caption{The poses, scenes, races, ages, glasses and expressions in LAMP-HQ.}
\vspace{0.2cm}
\scalebox{0.85}{
\begin{tabular}{c|ccccc}
\hline
\multirow{2}{*}{Poses} & \multicolumn{5}{l}{NIR: Pitch $\in {\left[\ang{0},\ang{10}\right]}$; Yaw $\in {\left[\ang{-45},\ang{45}\right]}$ }\\
 & \multicolumn{5}{l}{VIS: Pitch = $\ang{0}$; Yaw = $-\ang{45}$, $\ang{0}$, $\ang{45}$} \\
 \hline
\multirow{2}{*}{Scenes} & \multicolumn{5}{l}{\multirow{2}{*}{\begin{tabular}[c]{@{}l@{}}indoor: natural light, strong light, dim light, \\            outdoor: natural light, backlight\end{tabular}}} \\
 & \multicolumn{5}{l}{} \\
 \hline
\multirow{2}{*}{Races} & \multicolumn{5}{l}{\multirow{2}{*}{Asian, white, black}} \\
 & \multicolumn{5}{l}{} \\
 \hline
\multirow{2}{*}{Ages} & \multicolumn{5}{l}{\multirow{2}{*}{6-70}} \\
 & \multicolumn{5}{l}{} \\
 \hline
\multirow{2}{*}{Glasses Kinds} & \multicolumn{5}{l}{\multirow{2}{*}{\begin{tabular}[c]{@{}l@{}}sunglasses: 2, round glasses: 8,  square glasses: 5\end{tabular}}} \\
 & \multicolumn{5}{l}{} \\
 \hline
\multirow{2}{*}{Expressions} & \multicolumn{5}{l}{\multirow{2}{*}{neutral, closing eyes, smile}} \\
 & \multicolumn{5}{l}{}\\
 \hline
\end{tabular}
}
\end{table}

\subsection{Protocol}
To construct the uniform dataset partition, we provide a fair evaluation protocol for our LAMP-HQ. The subjects in our database are randomly divided into training set with 300 subjects and testing set with 273 subjects. For the training set, there are 29,525 NIR and 8,798 VIS images, and for the testing set, VIS images are used for the gallery set (only one frontal neutral image of each subject), while the NIR images, totally 17,163 images, are collected into the probe set. All images are saved in jpg format.

The typical usage of the protocol is to train a generative model on the training set. Then, a fixed VIS face recognition model is applied to match VIS images in the gallery set and translated VIS images from NIR images in the probe set. The rank1 recognition rate and verification rate are used to evaluate the final performance.

\begin{figure*}[tb]
\begin{center}
\includegraphics[width=0.8\linewidth]{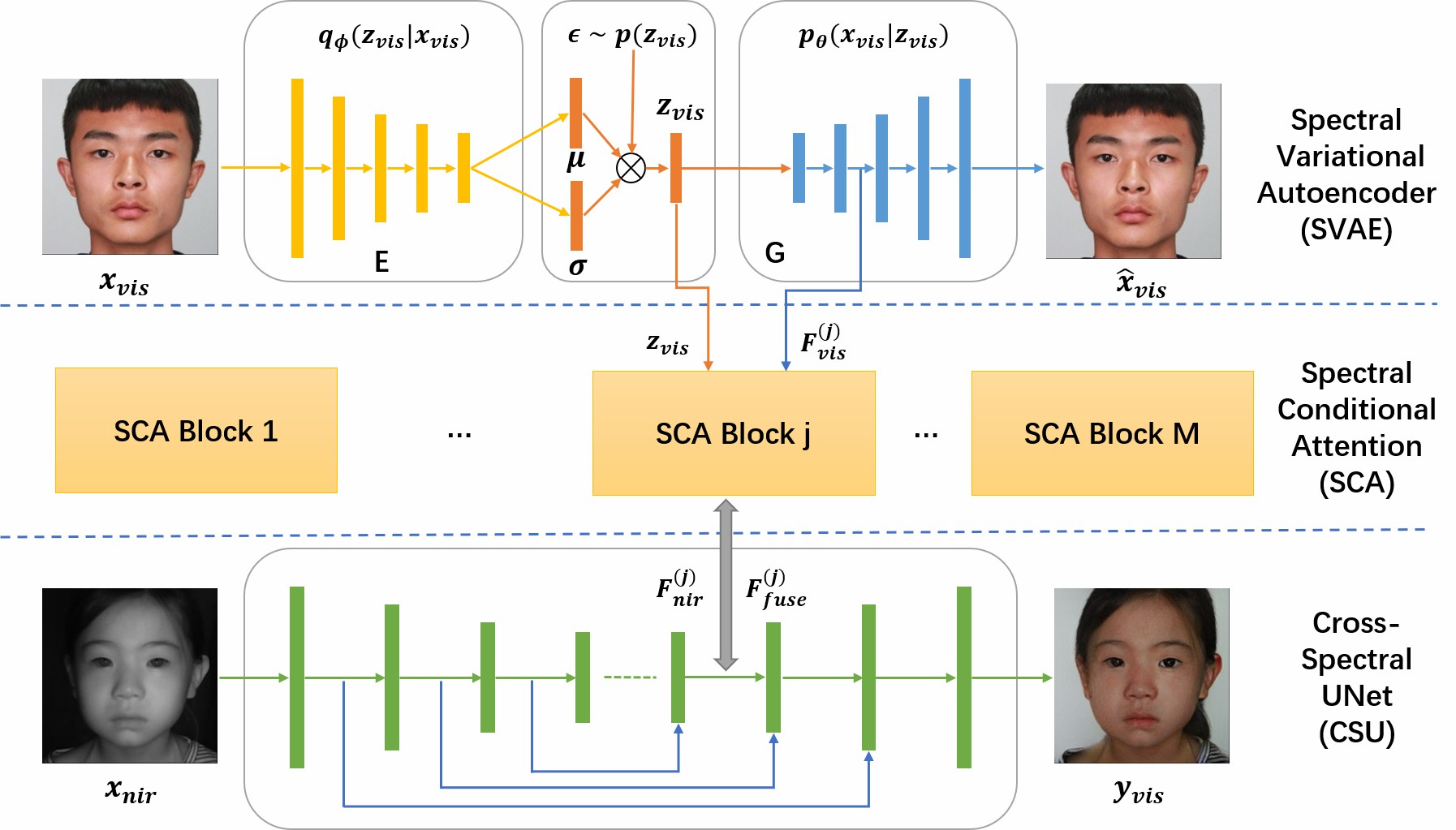}
\end{center}
   \caption{Overview of the proposed network architecture. It contains three parts: SVAE to extract the spectral representation $z_{vis}$, CSU to translate NIR data to the VIS domain, and SCA to transfer the VIS spectral style to CSU. Note that we only demonstrate the connections of one SCA block and neglect the others for a clear illustration.  }\label{fig:framework}
\end{figure*}

\section{Method}
In this section, we propose a variational spectral attention network (VSANet) for NIR-VIS image translation.
As illustrated in Fig.~\ref{fig:framework}, VSANet contains three parts: a spectral variational autoencoder (SVAE) to learn variational spectral representation $z_{vis}$ from the reference VIS image $x_{vis}$, a cross-spectral UNet (CSU) to translate the input NIR image $x_{nir}$ to its corresponding VIS version $y_{vis}$ along with the reference representation $z_{vis}$, and a spectral conditional attention (SCA) module to guide the combination of the content of $x_{nir}$ and the spectral style of $x_{vis}$.
The cross-spectral UNet  utilizes the same network architecture as the generator in CycleGAN~\citep{zhu2017unpaired}. The spectral variational autoencoder and spectral conditional attention module are elaborated together with the optimization objective functions in the following.

\begin{figure}[thb]
\begin{center}
\includegraphics[width=0.85\linewidth]{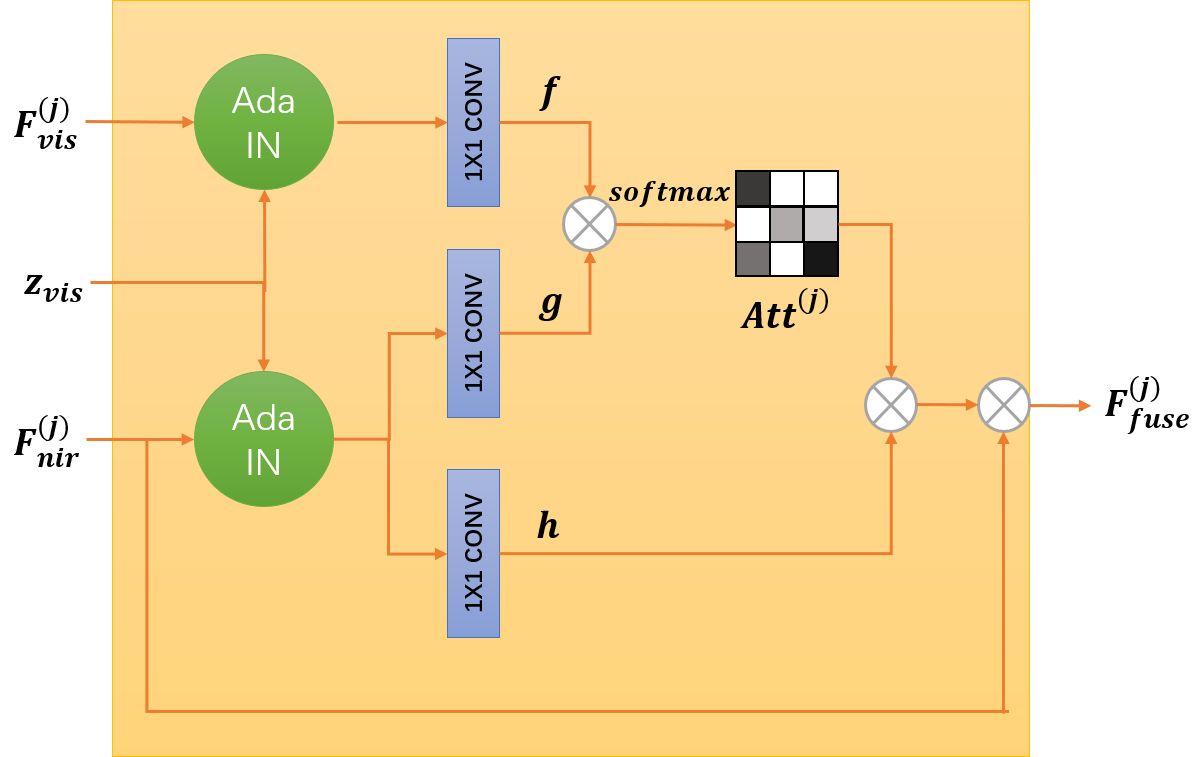}
\end{center}
   \caption{The spectral conditional attention (SCA) block. It contains two AdaIN operations and one conditional attention layer to transfer the spectral styles both globally and locally.
   }\label{fig:attention}
\end{figure}

\subsection{Spectral Variational Autoencoder}
Variational autoencoder (VAE)~\citep{kingma2013auto} is one of the most popular generative models that can learn precise manifold representations in an unsupervised way.
We design a spectral VAE (SVAE) to learn variational spectral representation $z_{vis}$ from the reference VIS image $x_{vis}$.
As shown in Fig.~\ref{fig:framework}, SVAE consists of two subnetworks: an inference network $E$ that maps VIS data $x_{vis}$ to the latent $z_{vis}$,  which approximates a prior \(p_{vis}(z_{vis})\), and a generative network $G$ that samples VIS data $\hat{x}_{vis}$ from $z_{vis}$.
The object of SVAE is to maximize the variational lower bound (or evidence lower bound, ELBO) of \(p_\theta (x_{vis})\):
\begin{equation}\label{eq:svae:elbo}
\begin{split}
  log p_\theta (x_{vis}) \geq & E_{q_\phi (z_{vis}|x_{vis})} \log p_\theta (x_{vis}|z_{vis}) \\
  & -  D_{KL} (q_\phi (z_{vis}|x_{vis}) || p(z_{vis})),
\end{split}
\end{equation}
where the first term on the right denotes the reconstruction accuracy for the output $\hat{x}_{vis}$, and the second regularizes the posterior $q_\phi (z_{vis}|x_{vis})$ to match the prior  $p(z_{vis})$.
Optimizing such ELBO, the spectral representation $z_{vis}$ can be sampled from either the posterior $q_\phi(z_{vis}|x_{vis})$ or the prior $p(z_{vis})$, which leads to the capability of the proposed method to translate NIR data to the VIS domain with or without VIS references.

Table 3 and Table 4 are the network architectures of the inference network $E$ and the generator $G$ in spectral VAE (SVAE), respectively. Given a VIS image of $3\times 256\times 256$, $E$ encodes it into two 512-d vectors, i.e., $\mu$ and $\sigma$, which forms the posterior $q_\phi(z_{vis}|x_{vis}) = \mathbb{N} (z;\mu, \sigma^2)$. The generator $G$ produces the reconstruction image $\hat{x}_{vis}$ from the spectral latent $z_{vis}$, where $z_{vis}\sim q_\phi(z_{vis}|x_{vis})$. Noted that we use $z_{vis}$ with two middle feature maps, i.e., $F_{vis}^{(3)}$ and $F_{vis}^{(4)}$ in Table 4, to guide the NIR-VIS translation.

\begin{table*}[htb]
\begin{center}
\label{tab:svae:e}

  \caption{Structure of the inference network $E$ of SVAE.}
    \begin{tabular}{|c|c|c|c|c|c|}

    \hline
    Input & Layer & Norm  & Act   & Output & Output Size \\
    \hline
    X$_{vis}$  & Conv3 & IN    & LeakyReLU & X0    & 32$\times$256$\times$256 \\

    X0    & MaxPool & /     & /     & X0    & 32$\times$128$\times$128 \\

    X0    & Conv3 & IN    & LeakyReLU & X0    & 64$\times$128$\times$128 \\

    X0    & MaxPool & /     & /     & X0    & 64$\times$64$\times$64 \\

    X0    & Conv3 & IN    & LeakyReLU & X0    & 128$\times$64$\times$64 \\

    X0    & MaxPool & /     & /     & X0    & 128$\times$32$\times$32 \\

    X0    & Conv3 & IN    & LeakyReLU & X0    & 256$\times$32$\times$32 \\

    X0    & MaxPool & /     & /     & X0    & 256$\times$16$\times$16 \\

    X0    & Conv3 & IN    & LeakyReLU & X0    & 512$\times$16$\times$16 \\

    X0    & MaxPool & /     & /     & X0    & 512$\times$8$\times$8 \\

    X0    & Conv3 & IN    & LeakyReLU & X0    & 512$\times$8$\times$8 \\

    X0    & FC & /     & /     & $\mu$, $\delta$    & 512, 512 \\
    \hline

    \end{tabular}\vspace{0.2cm}
    \end{center}
\end{table*}%

\begin{table*}[htb]
\label{tab:svae:g}
  \centering
  \caption{Structure of the generator network $G$ of SVAE.}
    \begin{tabular}{|c|c|c|c|c|c|}

    \hline
    Input & Layer & Norm  & Act   & Output & Output Size \\
    \hline
     {\color{red}z$_{vis}$}  & FC & / & / & F$_{vis}^{(1)}$ & 512$\times$8$\times$8 \\

    F$_{vis}^{(1)}$ & Upsample & /     & /     & F$_{vis}^{(1)}$ & 512$\times$16$\times$16 \\

    F$_{vis}^{(1)}$ & Conv3 & IN    & LeakyReLU & F$_{vis}^{(2)}$ & 512$\times$16$\times$16 \\

    F$_{vis}^{(2)}$ & Upsample & /     & /     & F$_{vis}^{(2)}$ & 512$\times$32$\times$32 \\

    F$_{vis}^{(2)}$ & Conv3 & IN    & LeakyReLU & {\color{red}F$_{vis}^{(3)}$} & 256$\times$32$\times$32\\

    F$_{vis}^{(3)}$ & Upsample & /     & /     & F$_{vis}^{(3)}$ & 256$\times$64$\times$64 \\

    F$_{vis}^{(3)}$ & Conv3 & IN    & LeakyReLU & {\color{red}F$_{vis}^{(4)}$} & 128$\times$64$\times$64 \\

    F$_{vis}^{(4)}$ & Upsample & /     & /     & F$_{vis}^{(4)}$ & 128$\times$128$\times$128 \\

    F$_{vis}^{(4)}$ & Conv3 & IN    & LeakyReLU & F$_{vis}^{(5)}$ & 64$\times$128$\times$128 \\

    F$_{vis}^{(5)}$ & Upsample & /     & /     & F$_{vis}^{(5)}$ & 64$\times$256$\times$256 \\

    F$_{vis}^{(5)}$ & Conv3 & IN    & LeakyReLU & F$_{vis}^{(6)}$ & 32$\times$256$\times$256 \\

    F$_{vis}^{(6)}$ & Conv3 & IN    & LeakyReLU & F$_{vis}^{(6)}$ & 3$\times$256$\times$256 \\

    F$_{vis}^{(6)}$ & /     & /     & sigmoid & $\hat{X}_{vis}$ & 3$\times$256$\times$256 \\
    \hline

    \end{tabular}%
    \vspace{0.2cm}
\end{table*}%

\subsection{Spectral Conditional Attention}

\begin{table}[htbp]
  \centering
  \caption{Structure of CSU injected with SCA blocks. The CBlock stands for a block of two convolutions, following by Instance Normalization and leakyReLU for each one.}
    \begin{tabular}{|c|c|c|c|}
    \hline

    Input & Layer &  Output & Output Size \\
    \hline
    X$_{nir}$  & C Block & F$_{nir}^{(1)}$ & 64$\times$256$\times$256 \\

    F$_{nir}^{(1)}$ & MaxPool &  F$_{nir}^{(2)}$ & 64$\times$128$\times$128 \\

    F$_{nir}^{(2)}$ & C Block &  F$_{nir}^{(2)}$ & 128$\times$128$\times$128 \\

    F$_{nir}^{(2)}$ & MaxPool & F$_{nir}^{(3)}$ & 128$\times$64$\times$64 \\

    F$_{nir}^{(3)}$ & C Block &  F$_{nir}^{(3)}$ & 256$\times$64$\times$64 \\

     {\color{red}F$_{nir}^{(3)}$, z$_{vis}$, F$_{vis}^{(3)}$} & {\color{red} SCA 1}  & {\color{red}F$_{fuse}^{(3)}$} & {\color{red}256$\times$64$\times$64} \\

    F$_{fuse}^{(3)}$ & MaxPool &  F$_{nir}^{(4)}$ & 256$\times$32$\times$32 \\

    F$_{nir}^{(4)}$ & C Block & F$_{nir}^{(4)}$ & 512$\times$32$\times$32 \\

    {\color{red}F$_{nir}^{(4)}$, z$_{vis}$, F$_{vis}^{(4)}$} & {\color{red}SCA 2}  & {\color{red} F$_{fuse}^{(4)}$} & {\color{red}512$\times$32$\times$32} \\

    F$_{fuse}^{(4)}$ & MaxPool & F$_{nir}^{(5)}$ & 512$\times$16$\times$16 \\

    F$_{nir}^{(5)}$ & C Block & F$_{nir}^{(5)}$ & 512$\times$16$\times$16 \\

    F$_{nir}^{(5)}$ & Upsample & F$_{nir}^{(5)}$ & 512$\times$32$\times$32 \\

    F$_{nir}^{(5)}$,F$_{nir}^{(4)}$ & concat & F$_{nir}^{(6)}$ & 1024$\times$32$\times$32 \\

    F$_{nir}^{(6)}$ & C Block &  F$_{nir}^{(6)}$ & 256$\times$32$\times$32 \\

    {\color{red}F$_{nir}^{(6)}$, z$_{vis}$, F$_{vis}^{(4)}$} & {\color{red}SCA 3}  &  {\color{red}F$_{fuse}^{(6)}$} & {\color{red}256$\times$32$\times$32} \\

    F$_{fuse}^{(6)}$ & Upsample &  F$_{fuse}^{(6)}$ & 256$\times$64$\times$64 \\

    F$_{fuse}^{(6)}$, F$_{nir}^{(3)}$ & concat &  F$_{nir}^{(7)}$ & 512$\times$64$\times$64 \\

    F$_{nir}^{(7)}$ & C Block &F$_{nir}^{(7)}$ & 128$\times$64$\times$64 \\

    {\color{red}F$_{nir}^{(7)}$, z$_{vis}$, F$_{vis}^{(3)}$}  & {\color{red}SCA 4}  &  {\color{red}F$_{fuse}^{(7)}$} & {\color{red}128$\times$64$\times$64} \\

    F$_{fuse}^{(7)}$ & Upsample &  F$_{fuse}^{(7)}$ & 128$\times$128$\times$128 \\

    F$_{fuse}^{(7)}$, F$_{nir}^{(2)}$ & concat & F$_{nir}^{(8)}$ & 256$\times$128$\times$128 \\

    F$_{nir}^{(8)}$ & C block & F$_{nir}^{(8)}$ & 64$\times$128$\times$128 \\

    F$_{nir}^{(8)}$ & Upsample & F$_{nir}^{(8)}$ & 64$\times$256$\times$256 \\

    F$_{nir}^{(8)}$, F$_{nir}^{(1)}$ & concat &  F$_{nir}^{(9)}$ & 128$\times$256$\times$256 \\

    F$_{nir}^{(9)}$ & C Block& F$_{nir}^{(9)}$ & 64$\times$256$\times$256 \\

    F$_{nir}^{(9)}$ & Conv3 & F$_{nir}^{(9)}$ & 3$\times$256$\times$256 \\

    F$_{nir}^{(9)}$ & Sigmoid     & Y$_{nir}$  & 3$\times$256$\times$256 \\

    \hline

    \end{tabular}%
    \vspace{0.2cm}

\end{table}%

As illustrated in Fig.~\ref{fig:framework}, a spectral conditional attention (SCA) module  is designed to build a bridge between the referenced spectral information learned by SVAE and the NIR-VIS translation flow in CSU. It consists of several multi-scale SCA blocks, each of which aims to produce such features that combine the VIS spectral style and the NIR content information. Then. the fused features are injected into the translation flow at the corresponding scale in CSU.

We employ two types of referenced VIS features to control the spectral style of the output $y_{vis}$ both globally and locally.
One is the spectral representation $z_{vis}\in \mathbb{R}^{C_z}$ sampled from $q_\phi(z_{vis}|x_{vis})$ or $p(z_{vis})$,
and the other is the feature $F^{(j)}_{vis}\in \mathbb{R}^{C_j\times H_j\times W_j}$ from the $i_j$th layer of $G$ in SVAE, i.e., $F^{(j)}_{vis} = F^{(i_j)}(z_{vis})$, where $j = 1,...,M$ denotes the index of the SCA block, $M$ is the number of SCA blocks. Given the content feature $F^{(j)}_{nir} \in \mathbb{R}^{C_j\times H_j\times W_j}$ from the $k_j$th layer of CSU, the fused feature $F^{(j)}_{fuse}$ can be obtained by the $j$th SCA block $B^{(j)}_{sca}$,
\begin{equation}\label{eq:sca}
  F^{(j)}_{fuse} = B^{(j)}_{sca} (F^{(j)}_{nir}, F^{(j)}_{vis}, z_{vis}).
\end{equation}
For $z_{vis}$ contains the whole information to generate a VIS image vis $G$ and $F^{(j)}_{vis}$ contains spatial style features of the size $H_j\times W_j$, they are expected to guide the global and local spectral styles, respectively.

As shown in Fig.~\ref{fig:attention}, the SCA block can be divided into two parts: adaptive instance normalization (AdaIN)~\citep{huang2017arbitrary} operation and conditional attention layer. $F^{(j)}_{nir}$ and $ F^{(j)}_{vis}$ are firstly re-normalized with the spectral $z_{vis}$ in AdaIN and then the attention map is computed to model the spatial similarities between the processed spectral feature $F^{'(j)}_{vis}$ and content feature $F^{'(j)}_{nir}$ in the conditional attention layer. The AdaIN operation is defined as
\begin{equation}\label{eq:adain}
  F^{'(j)}_{t}  = m^{(s)}(z_{vis}) \frac{F^{(j)}_{t} - \mu(F^{(j)}_{t})}{\sigma(F^{(j)}_{t})} + m^{(b)}(z_{vis}),
\end{equation}
where $t\in \{vis, nir\}$ and $m$ are an 8-layer MLP with two output vectors $m^{(s)}$ and $m^{(b)}$.

Inspired by the self-attention mechanism~\citep{zhang2019self}, we design a conditional attention layer to capture the spatial local relationships between the spectral and content features, i.e., $F^{'(j)}_{vis}$ and $F^{'(j)}_{nir}$. The attention map $Att^{(j)}\in \mathbb{R}^{(H_j W_j)\times (H_j W_j)}$ is obtained by
\begin{equation}\label{eq:att}
  Att^{(j)} =  softmax(f^{\mathrm{ T }}(F^{'(j)}_{vis})g(F^{'(j)}_{nir})),
\end{equation}
where $f$ and $g$ are $1\times 1$ convolutions.
The SCA block's output $F^{(j)}_{fuse}$  is computed by
\begin{equation}\label{eq:att}
  F^{(j)}_{fuse} =  F^{(j)}_{nir} + \gamma h(F^{'(j)}_{nir})Att^{(j)},
\end{equation}
where $h$ is a $1\times 1$ convolution and $\gamma$ is a learned parameter.

In this way, the SCA block can learn the spectral information by considering both the global and local spectral styles contained in $z_{vis}$ and $G$. 
Table 5 reports the network architecture of the cross-spectral UNet (CSU) injected with the spectral condition attention (SCA) blocks.
CSU produces VIS facial image $y_{nir}$ from the input NIR image $x_{nir}$ together with the spectral guidance $z_{vis}$, $F_{vis}^{(3)}$ and $F_{vis}^{(4)}$.  As shown in Tabel 5, we use four SCA blocks to fuse the exemplar spectral information with the features learned by CSU. The fused features are employed as the input for the next layer in CSU. 
The red color emphasizes the connections between  CSU and SCAs.

\subsection{Loss Functions}
In SVAE, the posterior $q_\phi(z_{vis}|x_{vis})$  is set to be a centered isotropic multivariate Gaussian $\mathbb{N}\left( z_{vis}; \mu_{vis}, \sigma_{vis}^2 \right)$, where $\mu_{vis}$ and $\sigma_{vis}$ are the output vectors of $E$. The prior $p(z_{vis})$ is set to be a simple Gaussian $\mathbb{N}\left( {{\bf{0}},{\bf{I}}}\right)$.  $z_{vis}$ for $G$
is sampled from $\mathbb{N}\left( z_{vis}; \mu_{vis}, \sigma_{vis}^2 \right)$ using a reparameterization trick, i.e., ${z_{vis}} = {\mu_{vis}} + \epsilon \odot \sigma_{vis}$, where $\epsilon \sim \mathbb{N}\left( {{\bf{0}},{\bf{I}}} \right)$.
The negative version of the two terms in Eq.~(\ref{eq:svae:elbo}) can be defined as
\begin{equation}\label{eq:svae:rec_kl}
\begin{split}
  L_{svae} &=\frac{1}{2} \| x_{vis} - \hat{x}_{vis} \|^{2}_F + \\
            & \frac{1}{2}  \sum_{i=1}^{c_z}  ( (\mu_{vis}^{i})^2 + (\sigma_{vis}^{i})^2 - \log ((\sigma_{vis}^{i})^2) - 1).
\end{split}
\end{equation}

For the optimization of CSU and SCA, we utilize four losses, i.e., the content loss $L_{content}$, the style loss $L_{style}$, the identity-preserving loss $L_{id}$ and the adversarial loss $L_{adv}$, to produce high-fidelity VIS facial images. Among them, $L_{content}$ and $L_{style}$ are employed to combine the content of NIR data and the style of VIS data, respectively; $L_{id}$ is used to preserve the identity information and $L_{adv}$ is used to improve the image quality of the output image. The details are given below.

Similar to~\citep{johnson2016perceptual}, we employ the VGG-16 network ~\citep{simonyan2014very} to compute the content loss to restrict the content similarity between the input $x_{nir}$ and the output $y_{vis}$. The content loss is obtained by computing the Euclidean distance between the features of $x_{vis}$ and $y_{vis}$ extracted by VGG-16. It can be formulated as follows:
\begin{equation}\label{eq:content}
  L_{content} =\frac{1}{C_{j}H_{j}W_{j}} \| U_{j}(y_{vis}) - U_{j}({x}_{nir}) \|_{2}^{2}
\end{equation}
where $U_{j}(y_{vis})$ and $U_{j}({x}_{nir})$ are the ouput features of the $j$th layer of VGG-16, repsectively. We use the relu3\_3 layer to extract the features for the content loss.

In order to preserve the spectral information learned from $x_{vis}$ in the process of generating the output $y_{vis}$, we use the style loss proposed in~\citep{gatys2016image}. It is obtained by computing the distance between the Gram matrices of $x_{vis}$ and $y_{vis}$, which is defined as:
\begin{equation}
    L_{style} = \sum \limits_{j} \| G^{U}_{j}(x_{vis}) - G^{U}_{j}({y}_{nir}) \|^{2}_{F},
\end{equation}
where
\begin{equation}
    G^{U}_{j}(x)_{c,c'} = \\ \frac{1}{C_{j}H_{j}W_{j}} \sum_{h=1}^{H_j} \sum_{w=1}^{W_j} U_{j}(x)_{h,w,c}U_{j}(x)_{h,w,c'},
\end{equation}
$U_{j}(x)$ is the output feature of the $j$th layer of VGG-16. We use the relu1\_2, relu2\_2, relu3\_3 and relu4\_3 layers to extract the features for the style loss.

As discussed before, LAMP-HQ contains facial images of different poses for each subject. In the training phase, we introduce a frontal VIS image $x_{vis}^{(match)}$ that has the same identity with the input NIR $x_{nir}$. The produced image $y_{vis}$ is expected to have the most identity characteristics with
$x_{vis}^{(match)}$. For this purpose, an identity-preserving loss is designed in a similar way with the content loss, which is formulated as:
\begin{equation}
  L_{id} = \|F(x_{vis}^{(match)}) - F(y_{vis})\|_{1},
\end{equation}
where $F(x_{vis}^{(match)})$ and $F(y_{vis})$ are the extracted features respectively for $x_{vis}^{(match)}$ and $y_{vis}$ by a face recognition network, i.e., a pretrained LightCNN in this paper. We utilize the feature before the last FC layer of LightCNN for the indentity-preserving loss.

Generative Adversarial Network (GAN)~\citep{goodfellow2014generative}
is one of the most popular generative models nowadays. It can produce photo-realistic images via playing a min-max game between a generator network and a discriminator network. We utilize GAN to improve the visual quality of the produced VIS image $y_{vis}$, where the proposed CSU serves as the generator $G$ of GAN.
The two-player minimax game is,
\begin{equation}
\begin{split}
    \min \limits_{G}\max \limits_{D} V(D, G) = \mathbb{E}_{x\sim p_{vis}(x)}\left [ \log D(x) \right ] + \\ \mathbb{E}_{y_{vis}\sim p_{g}(y_{vis})}\left [ \log (1 - D(y_{vis}) )\right ]
\end{split}
\end{equation}
where $p_{vis}(x)$ represents the real distribution of VIS data and $p_{g}(y_{vis})$ represents the distribution of the generated VIS data. Then the adversarial loss can be formulated as,
\begin{equation}
    L_{adv} = -\log(D(y_{vis})))
\end{equation}

The total loss is defined as
\begin{equation}\label{eq:svae:total}
\begin{split}
  L_{total} = L_{content} + \lambda_1 L_{style} + \lambda_2 L_{id} +\lambda_3  L_{adv},
\end{split}
\end{equation}
where $\lambda_1$, $\lambda_2$ and $\lambda_3$ are trade-off parameters.

\subsection{Network Architecture}
Table 6 detailedly reports the network architecture of the discriminator $D$ used in the computation of the adversarial loss. The discriminator $D$ and and the CSU network along with the SCA module are optimized iteratively following the original GAN~\citep{goodfellow2014generative}.

\begin{table*}[htbp]
  \centering
  \caption{Structure of the discriminator. }
    \begin{tabular}{|c|c|c|c|c|c|}

\hline
    Input & Layer & Norm  & Act   & Output & Output Size \\
    \hline
    X$_{vis}$/Y$_{vis}$ & Conv4 & /     & LeakyReLU & X0    & 64$\times$128$\times$128 \\

    X0    & Conv4 & IN    & LeakyReLU & X0    & 128$\times$64$\times$64 \\

    X0    & Conv4 & IN    & LeakyReLU & X0    & 256$\times$32$\times$32 \\

    X0    & Conv4 & /     & /     & out    & 1$\times$16$\times$16 \\
    \hline

    \end{tabular}%
    \vspace{0.2cm}

\end{table*}%

\section{Experiments}

We evaluate our method qualitatively and quantitatively on the proposed LAMP-HQ database. For qualitative evaluation, we show the results of synthetic VIS images from corresponding input NIR images. For quantitative evaluation, we perform cross-spectral face recognition based on original and synthesized face images. We also provide three HFR benchmarks on LAMP-HQ, including Pixel2Pixel~\citep{isola2017image}, CycleGAN~\citep{zhu2017unpaired} and ADFL~\citep{song2018adversarial}. Both LightCNN-9 and LightCNN-29~\citep{wu2018light} are employed as face classifiers in the experiment. To further demonstrate the effectiveness of our method and assess the difficulty of LAMP-HQ, we also conduct experiments on CASIA NIR-VIS 2.0 Face Database~\citep{li2013casia}, BUAA-VisNir face database~\citep{huang2012BUAA} and Oulu-CASIA NIR-VIS database~\citep{chen2009learning}, which are widely used in the HFR field.

We crop and then align the facial images of pixel-size $256\times 256$ on all the above databases. Adam optimizer is employed with a learning rate of 2e-4, the $\beta_1$ of 0.5 and the $\beta_2$ of 0.99. The batch-size is set to $8$ and the model converges in approximately 60,000 iterations. The trade-off parameters $\lambda_1$, $\lambda_2$, and $\lambda_3$ in Eq.~(\ref{eq:svae:total}) are
set empirically to be $1$, $5$ and $0.1$, respectively.

\begin{figure}[htbp]
\begin{center}
\includegraphics[width=0.8\linewidth]{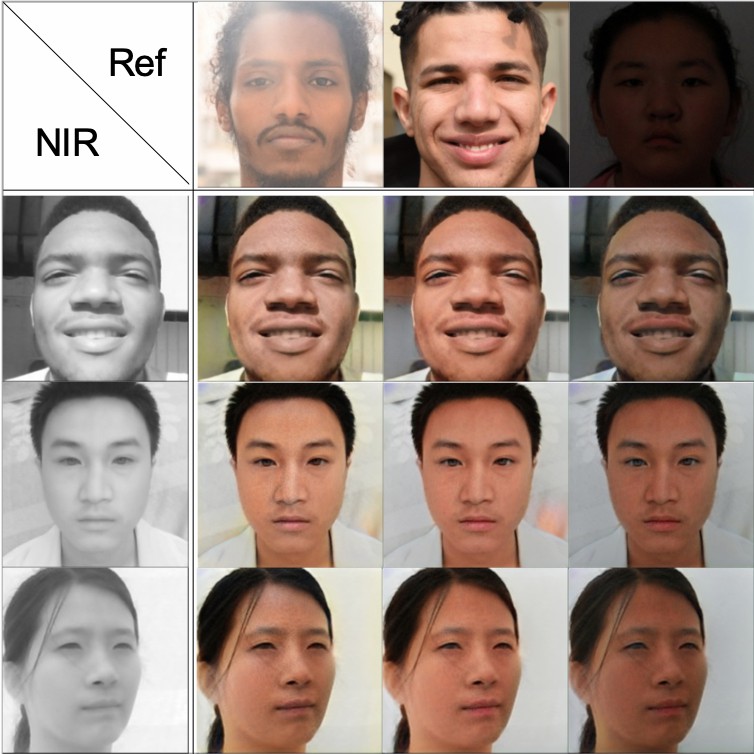}
\end{center}
   \caption{The qualitative results of exemplar-based NIR-VIS translation. Three spectral styles of VIS data in the top row are transferred to three NIR images in the left column. Nine generated images are in the middle.}\label{fig:visual:ref}
\end{figure}

\begin{figure*}[htbp]
\begin{center}
\includegraphics[width=0.9\linewidth]{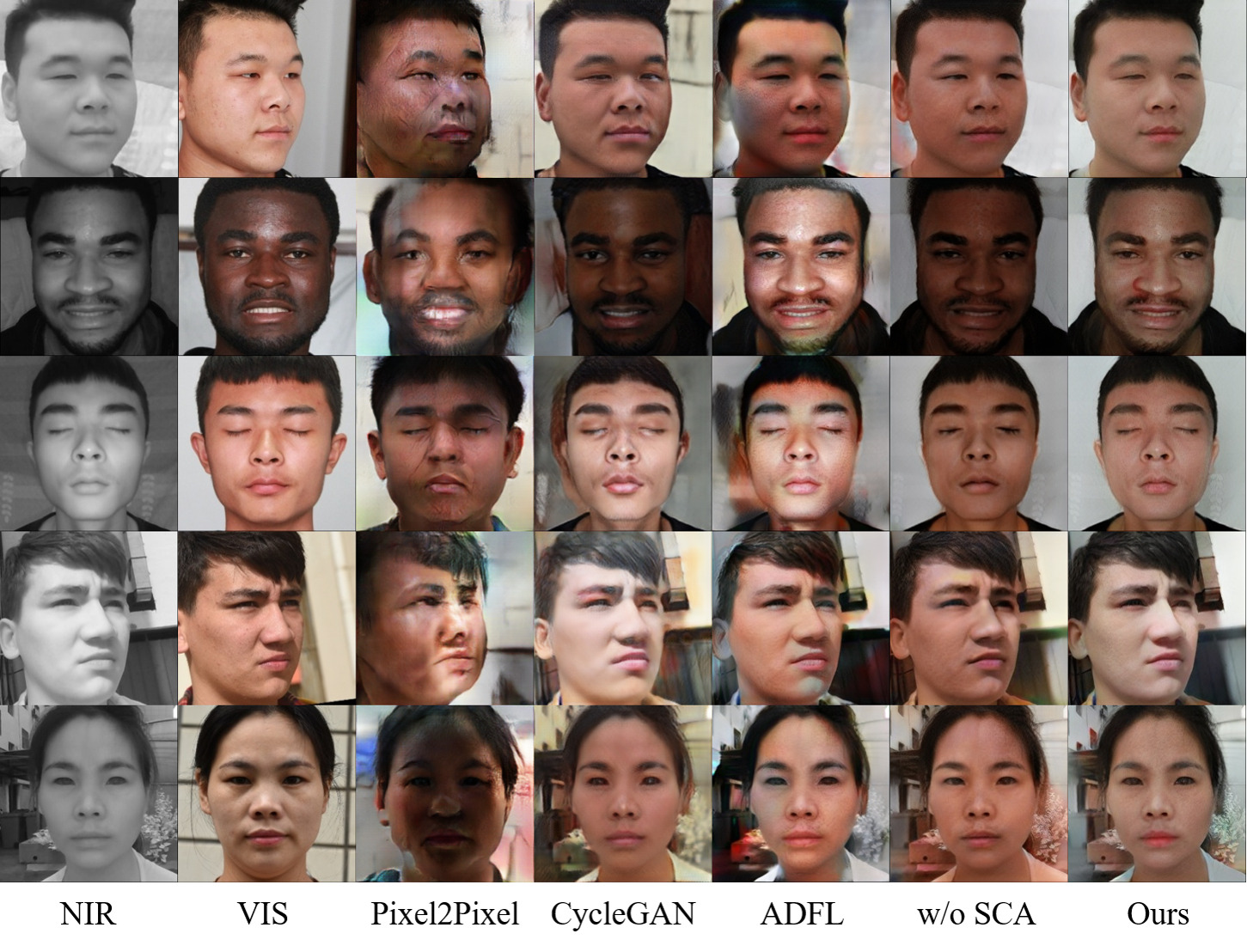}
\end{center}
   \caption{Qualitative comparison on the LAMP-HQ dataset. From left to right, the columns are the input NIR data, the corresponding VIS data, the results of Pixel2Pixel~\citep{isola2017image}, CycleGAN~\citep{zhu2017unpaired}, ADFL~\citep{song2018adversarial}, our method without SCA, and our method, respectively. }\label{fig:visual:results}
\end{figure*}

\begin{table*}[htb]
\centering
\label{tab:comp:lamp:quan}
\caption{NIR-VIS face recognition on LAMP-HQ.}
\vspace{0.2cm}
\scalebox{1}{
\begin{tabular}{l|ccc|ccc}
\hline
\multirow{2}{*}{Method} & \multicolumn{3}{c|}{LightCNN-9} & \multicolumn{3}{c}{LightCNN-29} \\
                        & Rank-1  & FAR=1\%  & FAR=0.1\%  & Rank-1   & FAR=1\%  & FAR=0.1\%  \\
\hline
original             &  89.77       &     88.46    &      71.61      & 94.94    & 93.42    & 78.65      \\
Pixel2Pixel             &  17.04       &  24.45        &    7.95        & 21.47    & 27.35    & 8.48       \\
cycleGAN                &   80.17      &  78.13        &   32.29         & 87.29    & 84.85    & 59.92      \\
ADFL                    &     88.09    &   87.67       &   68.06         & 93.65    & 92.35    & 74.95      \\
\hline
ours w/o $L_{content}$             &  54.21       &   53.53       &     29.05       & 63.49    & 60.62    & 37.55      \\
ours w/o $L_{style}$             &    87.98     &   86.38       &     65.42       & 94.35    & 92.57    & 74.03      \\
ours w/o $Lid$           &     86.84    &   86.97       &    64.91        & 94.12    & 92.75    & 74.03      \\
ours w/o $L_{adv}$           &    88.53     &    88.07      &     68.18       & 95.52    & 94.40    & 79.52      \\
ours w/o  AdaIN         &    93.67     &     91.22     &    72.89        & 96.70    & 95.05    & 80.78      \\
ours w/o  SCA           &     92.74    &    90.41      &    72.50        & 95.72    & 94.25    & 77.26      \\
\bf{ours}                    &      \bf{94.09}   &    \bf{91.81}      &    \bf{74.77}        & \bf{97.46}    & \bf{95.59}    & \bf{82.32}   \\
\hline
\bf{fine-tune 1}&/&/&/&\bf{94.23}&\bf{93.99}&\bf{80.74}\\
\hline
\bf{fine-tune 2}&/&/&/&\bf{95.81}&\bf{92.35}&\bf{81.46}\\
\hline
\end{tabular}
}
\end{table*}

\begin{table*}[htb]
\centering
\label{tab:quan:comp:other}
\caption{Quantitative comparisons on CASIA 2.0, BUAA, and Oulu NIR-VIS databases. The results of the compared methods are copied from the published papers.}
\vspace{0.2cm}
\scalebox{0.8}{
\begin{tabular}{l|ccc|ccc|ccc}
\hline
\multirow{2}{*}{Method} & \multicolumn{3}{c|}{CASIA 2.0} & \multicolumn{3}{c|}{BUAA}     & \multicolumn{3}{c}{Oulu}     \\
                        & Rank-1      & FAR=1\%    & FAR=0.1\%  & Rank-1 & FAR=1\% & FAR=0.1\% & Rank-1 & FAR=1\% & FAR=0.1\% \\
\hline
original                & 96.84       & 99.10      & 94.68      & 96.5   & 95.4    & 86.7      & 96.7   & 92.4    & 65.1      \\
H2(LBP3)                & 43.8        & 36.5       & 10.1       & 88.8   & 88.8    & 73.4      & 70.8   & 62.0    & 33.6      \\
TRIVET                  & $95.7\pm0.52$   & $98.1\pm0.31$  & $91.0\pm1.26$  & 93.9   & 93.0    & 80.9      & 92.2   & 67.9    & 33.6      \\
IDR                     & $97.3\pm0.43$   & $98.9\pm0.29$  & $95.7\pm0.73$  & 94.3   & 93.4    & 84.7      & 94.3   & 73.4    & 46.2      \\
Pixel2Pixel             & 22.13       & 39.22      & 14.45      & /      & /       & /         & /      & /       & /         \\
cycleGAN                & 87.23       & 93.92      & 79.41      & /      & /       & /         & /      & /       & /         \\
ADFL                    & $98.2\pm0.34$   & $99.1\pm0.15$  & $97.2\pm0.48$  & 95.2   & 95.3    & 88.0      & 95.5   & 83.0    & 60.7      \\
\hline
\bf{ours}                    & \bf{99.0}           & \bf{99.9}          & \bf98.3         & \bf98.0   & \bf98.2    & \bf92.5      & \bf99.9   & \bf96.8    & \bf82.3  \\
\hline
\end{tabular}
}
\end{table*}




\subsection{Qualitative Evaluation}
We conduct two types of NIR-VIS translation experiments on LAMP-HQ. One type produces VIS data by combining the content of NIR data and the spectral style of VIS exemplar, where the spectral representation $z_{vis}$ is sampled from the posterior $q_\phi(z_{vis}|x_{vis})$.
It can be observed in Fig.~\ref{fig:visual:ref} that given different VIS exemplars, the outputs of the same NIR input image have different spectral styles. This phenomenon verifies that the proposed method can extract spectral information from VIS data and transfers it to NIR data to produce photo-realistic VIS images.

The other type of NIR-VIS translation is generating VIS data from NIR data without VIS exemplar, where  $z_{vis}$ is sampled from the prior $p(z_{vis}) = \mathbb{N}(0,I)$. We compare the performance with other SOTA methods in this case. As demonstrated in Fig.~\ref{fig:visual:results}, the proposed method significantly outperforms other methods.
Pixel2Pixel fails to generate visual realistic images due to the lack of paired NIR-VIS data. CycleGAN produces artifacts in some hard cases, such as closing eyes or large poses. ADFL cannot reconstruct realistic textures in the background. The ablation version of our method without spectral conditional attention also fails to generate realistic colors and textures for the background. This demonstrates that spectral conditional attention mechanism is helpful for generating realistic local texture details. The comparison results verify the effectiveness of the proposed method for cross-spectral facial hallucination, even in extreme conditions.


\subsection{NIR-VIS Face Recognition}

Following the protocol described in Section 3, we conduct quantitative comparison  experiments on LAMP-HQ with several deep learning methods, including Pixel2Pixel, CycleGAN and ADFL.
The pre-trained models, i.e., LightCNN-9 and LightCNN-29, are utilized to compute three metrics i.e.,  Rank-1 accuracy, and verification rates when VR@FAR=1\%, 0.1\%, for evaluation. The results are reported in Table 7.
As shown in  Table 7,  the proposed method achieves the best performance of NIR-VIS face recognition. Compared to the original NIR data in the probe, the synthesized VIS data can improve Rank-1 accuracy from 89.77\% to 94.09\% by LightCNN-9 and  from 94.94\% to 97.46\% by LightCNN-29. This verifies that the proposed method is effective in improving the performance of NIR-VIS face recognition.

Besides, we fine-tune the LightCNN-29 on the proposed dataset and the results are shown in the last two lines of Table 7. The fine-tune $1$ and fine-tune $2$ stand for the fine-tuning on the training NIR-VIS images and the synthesized VIS images, respectively. The data show that the results decline a little than before. The reason may be that the NIR-VIS database is small-scale compared to the training dataset employed by LightCNN, which has millions images of more than one hundred thousand people~\citep{wu2018light}. Therefore the priori knowledge learned before is easily destroyed even after fine-tuning, then influences the recognition results.



\subsection{Evaluation on Other Databases}

In this section, we report the evaluation results on the CASIA 2.0, Oulu and BUAA databases.
For the CASIA 2.0 database, we use the standard protocol in View 1 for evaluation. For the BUAA  and Oulu databases, our model trained on LAMP-HQ is directly used to evaluate the testing sets of BUAA and Oulu following the standard protocol.
We compare the recently proposed H2(LBP3)~\citep{shao2016cross}, TRIVET~\citep{liu2016transferring}, IDR~\citep{he2017learning}, Pixel2Pixel~\citep{isola2017image}, CycleGAN~\citep{zhu2017unpaired}, and ADFL~\citep{song2018adversarial}.
The quantitative results of face recognition computed by LightCNN-29 are reported in Table 8.
We also design cross experiments of the proposed method on the $4$ databases above, the evaluation results are showed in Table 9.
It can be observed that the proposed method outperforms others on all  three databases.
Note that the results of our method on Oulu and BUAA are obtained by the model trained on LAMP-HQ,
which verifies the generalization of the proposed method.

Table 7 and Table 8 show that the performance of NIR-VIS face recognition on the proposed LAMP-HQ database is much lower than that on the CASIA 2.0 database. This indicates that LAMP-HQ is a more challenging NIR-VIS database for  face recognition.

\begin{table*}
\centering
\caption{Cross experiment results on Oulu, BUAA, CASIA2.0, and LAMP.}
\begin{tabular}{|c|ccccc|}
\hline
\diagbox{Train}{Test}& Oulu & BUAA & CASIA 2.0 & LAMP & Metric\\ 
\hline
\multirow{3}{*}{Oulu}&86.3&76.9&67.8&62.3&Rank-1\\
  & 71.5&72.1&83.1&79.3&FAR=1\%\\
  & 35.9&48.2&52.8&47.4&FAR=0.1\%\\
\hline
\multirow{3}{*}{BUAA}&97.1&97.8&92.5&90.9&Rank-1\\
  & 80.6&97.2&97.9&81.4&FAR=1\%\\
  & 50.6&92.7&90.6&76.2&FAR=0.1\%\\
\hline
 \multirow{3}{*}{CASIA 2.0}&98.5&96.4&98.0&94.5&Rank-1\\
  & 94.5&95.6&98.9&92.4&FAR=1\%\\
  &75.1&86.3&94.0&90.2&FAR=0.1\%\\
\hline
\multirow{3}{*}{LAMP}&99.9&98.0&99.0&97.5&Rank-1\\
  & 96.8&96.88&99.9&95.6&FAR=1\%\\
  &82.3&92.5&98.3&82.3&FAR=0.1\%\\
\hline
\end{tabular}
\label{tab:cross}
\end{table*}

\subsection{Ablation Study}

\begin{figure*}[htbp]
\begin{center}
\includegraphics[width=0.9\linewidth]{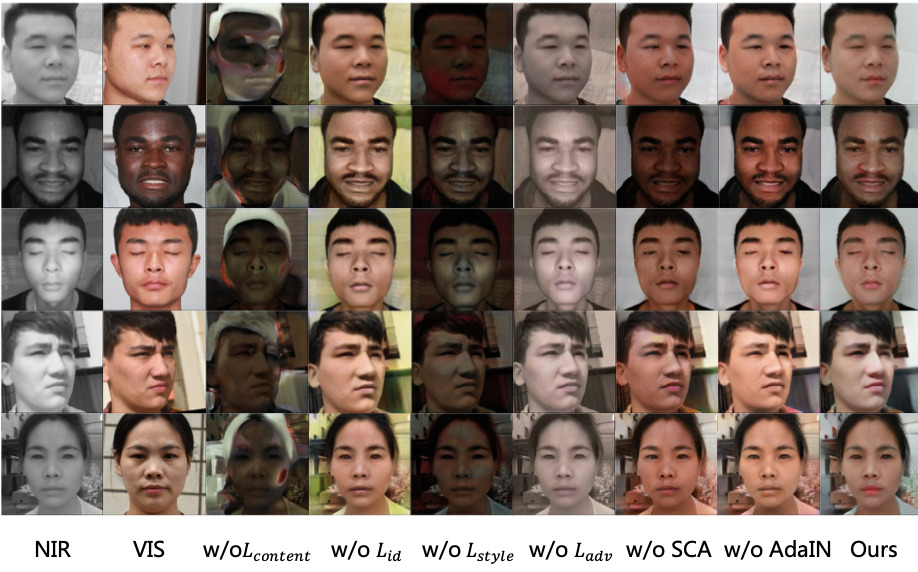}
\end{center}
   \caption{The visual results of the ablation study. From left to right, the columns are the input NIR data, the corresponding VIS data, the results of several version of the proposed method, i.e., the ablation versions without $L_{content}$, $L_{id}$, $L_{style}$, $L_{adv}$, the SCA module and the AdaIN operation, respectively. The last column are the results of the proposed method. }\label{fig:ablacation}
\end{figure*}

To indicate the effects of each component of our method on the performance of HFR, we show in Table 7 the results of the ablation study.
We can observe that the content loss in Eq.~(\ref{eq:svae:total}) is essential to preserve the input's content, including identity information.
The style loss, identity-preserving loss and adversarial loss
seem to play a similar role in boosting the recognition performance because all of them regularize the generated data to preserve certain VIS characteristics. When the AdaIN operation or the SCA module is absent, the recognition performance drops marginally. Fig.~\ref{fig:ablacation} shows the detailed visualization results.

\section{Conclusion}

This paper proposes a  large-scale multi-pose high-quality database for NIR-VIS heterogeneous face recognition. To the best of our knowledge, LAMP-HQ is the largest NIR-VIS database containing different illuminations, scenes, expressions, poses, and accessories. We also provide an efficient benchmark for NIR-VIS face recognition on LAMP-HQ, including Pixel2Pixel, CycleGAN, and ADFL. In addition, we propose a novel exemplar-based variational spectral attention network (VSANet) to combine the learned spectral information of referenced VIS images and the content information of input NIR images. In this way, a photo-realistic image can be generated that is helpful for cross-spectral face recognition. We hope that our LAMP-HQ database and the benchmark could make for the development of NIR-VIS face recognition.


\begin{acknowledgements}
This work is funded by Beijing Natural Science Foundation (Grants No. JQ18017).
\end{acknowledgements}

%
%

\bibliographystyle{spbasic}      
\bibliography{reference}   


\end{document}